\documentclass[journal]{IEEEtran}
\usepackage{amsmath,amsfonts}
\usepackage{algorithmic}
\usepackage{array}
\usepackage[caption=false,font=normalsize,labelfont=sf,textfont=sf]{subfig}
\usepackage{textcomp}
\usepackage{stfloats}
\usepackage{url}
\usepackage{verbatim}
\usepackage{graphicx}
\def\BibTeX{{\rm B\kern-.05em{\sc i\kern-.025em b}\kern-.08em
    T\kern-.1667em\lower.7ex\hbox{E}\kern-.125emX}}
\usepackage{balance}

\usepackage{amssymb,amsthm,bm,bbm,graphicx,color}
\usepackage[colorlinks, linkcolor=blue, anchorcolor=blue, citecolor=blue]{hyperref}
\usepackage{multirow}

\usepackage{algorithm}

\usepackage{colortbl}
\definecolor{mygray}{gray}{.9}
\definecolor{lightgray}{gray}{.7}

\newcommand{\mbf}[1]{\mathbf{#1}}

\begin{document}
\title{Learning Multi-Task Gaussian Process Over Heterogeneous Input Domains}

\author{Haitao~Liu,
	Kai~Wu,
	Yew-Soon~Ong,~\IEEEmembership{Fellow,~IEEE,}
	Chao~Bian,
	Xiaomo~Jiang
	and~Xiaofang~Wang
	\thanks{This work has been submitted to the IEEE for possible publication. Copyright may be transferred without notice, after which this version may no longer be accessible.}
	\thanks{This work was supported by the National Natural Science Foundation of China (52005074), and the Fundamental Research Funds for the Central Universities (DUT19RC(3)070).}
	\thanks{Haitao Liu, Kai Wu, Chao Bian and Xiaofang Wang are with School of Energy and Power Engineering, Dalian University of Technology, China, 116024. E-mail: htliu@dlut.edu.cn, wukai0048@mail.dlut.edu.cn, bc2019@mail.dlut.edu.cn, dlwxf@dlut.edu.cn. Xiaomo Jiang is with School of Energy and Power Engineering, and Digital Twin Laboratory for Industrial Equipment at Dalian University of Technology, China, 116024. E-mail: xiaomojiang2019@dlut.edu.cn. Yew-Soon Ong is with School of Computer Science and Engineering, Nanyang Technological University, Singapore, 639798. E-mail: asysong@ntu.edu.sg.}
}


\maketitle

\begin{abstract}
Multi-task Gaussian process (MTGP) is a well-known non-parametric Bayesian model for learning correlated tasks effectively by transferring knowledge across tasks. But current MTGPs are usually limited to the multi-task scenario defined in the same input domain, leaving no space for tackling the heterogeneous case, i.e., the features of input domains vary over tasks. To this end, this paper presents a novel heterogeneous stochastic variational linear model of coregionalization (\texttt{HSVLMC}) model for simultaneously learning the tasks with varied input domains. Particularly, we develop the stochastic variational framework with Bayesian calibration that (i) takes into account the effect of dimensionality reduction raised by domain mappings in order to achieve effective input alignment; and (ii) employs a residual modeling strategy to leverage the inductive bias brought by prior domain mappings for better model inference. Finally, the superiority of the proposed model against existing LMC models has been extensively verified on diverse heterogeneous multi-task cases and a practical multi-fidelity steam turbine exhaust problem.
\end{abstract}


\section{Introduction}
%
%
%
%
\IEEEPARstart{M}{ulti-task} Gaussian process (MTGP), also known as multi-output or multi-fidelity Gaussian process, has been developed and studied over decades~\cite{alvarez2012kernels, liu2018remarks}. Different from the conventional single-task GP~\cite{barbosa2021lateral, petrovic2022mixtures, nguyen2018efficient}, the MTGP particularly exploits and represents the correlations among related tasks in order to achieve knowledge sharing and transfer, which therefore improves the quality of prediction and alleviates the demand of large-scale training data. Consequently, the MTGPs have gained widespread application in diverse domains, for example, time series forecasting~\cite{durichen2014multitask}, multi-task optimization~\cite{swersky2013multi, wang2021choose} and multi-fidelity classification~\cite{klyuchnikov2020gaussian}.

As the representative of kernel method, given $T$ correlated tasks as well as the associated training data $\mathcal{D}=\{\mathcal{D}^t\}_{t=1}^T$, the key in MTGP is to build multi-task modeling framework as well as the related multi-task kernel. For instance, the well-known linear model of coregionalization (LMC) and the variants~\cite{bonilla2007multi, nguyen2014collaborative, ashman2020sparse, liu2021scalable}, which are the focus of this paper, utilize $Q$ shared latent GPs and linearly mix them to express the related $T$ tasks simultaneously. Differently, the convolved GP~\cite{alvarez2008sparse} expresses each task as the convolution of a smoothing and task-specific kernel with a common base process, thus resulting into a non-separable multi-task generative model. Alternatively, the exploitation of spectral mixture kernel has further improved model capability, see~\cite{parra2017spectral, chen2019multioutput}. Particularly, in order to account for the asymmetric scenario raised from for example multi-fidelity modeling~\cite{fernndez-godino2016review} and transfer learning~\cite{zhuang2021atl}, the asymmetric modeling structure has been investigated in the MTGP paradigm~\cite{le2014recursive, kandemir2015asymmetric, liu2018cope, requeima2019gaussian}. The MTGPs have also been studied in other regimes, like transfer learning~\cite{papevz2021transferring} and few-shot learning~\cite{patacchiola2019deep}.

Though successful stories have been successively reported in literature, the two disadvantages of current MTGPs however are (i) the poor scalability for handling massive data collected from multiple tasks, and (ii) the requirement of same input domain for all the tasks. As for the first issue, it is inherited from the full GP paradigm and becomes more serious for multi-task learning. It is known that given $N$ training points, the GP suffers from the cubic time complexity $\mathcal{O}(N^3)$ due to the operations on the $N \times N$ covariance matrix~\cite{liu2020gaussian}. As for MTGP, given $T$ tasks with $TN$ training points in total, the time complexity quickly increases to $\mathcal{O}(T^3N^3)$, which makes the training of model for many tasks and training points infeasible in practice. To alleviate this issue, the scalability of MTGP could be improved by leveraging the ideas developed in the paradigm of scalable GP~\cite{chiplunkar2016approximate}. The commonly used scalable strategies include (i) sparse approximation~\cite{titsias2009variational, hensman2013gaussian, liu2021scalableG} which introduces $M$ inducing random variables ($M \ll N$) to approximate the stochastic behavior, thus resulting in the time complexity reduced to $\mathcal{O}(NM^2)$ and even $\mathcal{O}(M^3)$ when using stochastic variational inference; and (ii) distributed approximation~\cite{liu2018generalized} which partitions the large-scale training data and aggregates predictions from local GP experts with low time complexity. Particularly, for the multi-task case with a large $T$, for example, modeling the data collected from many sensors in the physical field, it greatly increases the model complexity from another dimension. Recently, the dimensionality reduction and tensor decomposition have been exploited in the MTGP paradigm to make it scalable for many tasks~\cite{perdikaris2016multifidelity, yu2017tensor, zhe2019scalable, wang2020multi}.

As for the second issue, some pioneer works have raised in literature to tackle input domain alignment for multi-task/multi-fidelity modeling. When there exists prior knowledge extracted from expert opinion, for example, the prior domain mappings, we could directly incorporate the inductive bias to achieve input alignment and thereby use the conventional MTGP on the transformed domain, see for example~\cite{li2016integrating}. More flexibly, in order to achieve model calibration, Tao et al.~\cite{tao2019input} proposed to learn a linear embedding to align two tasks with different input domains. Furthermore, given the aligned inputs known at training points, Hebbal et al.~\cite{hebbal2021multi} decided to introduce another GP to model the input transformation. Besides, some ideas developed from other communities may help address this issue. For example, the transfer learning community leverages the maximum mean discrepancy (MMD) criterion to measure the similarity of data distributions from different domains in a high-dimensional latent space~\cite{long2017deep}, thus maximizing the MMD criterion could make the domains close to each other. In addition, the multi-view GPs~\cite{li2018shared, mao2020multiview} have been proposed to achieve input alignment through for example shared mean of inducing variables, which however targets on modeling the same task from multiple views (data). Finally, it is worth noting that except for the heterogeneousness raised in input domain, the output (task) domain may also have difference, for example, the mixture of regression and binary classification considered in~\cite{moreno-muoz2018heterogeneous}. 

To address the above two issues, especially the second one, this article presents a heterogeneous stochastic variational LMC model with \textit{Bayesian calibration} for multi-task learning on varied input domains. Specifically, the main contributions of this paper are three-fold:
\begin{itemize}
	\item A Bayesian calibration method for input alignment as well as the related stochastic variational modeling framework has been presented for heterogeneous MTGP. It particularly takes into account the effect of dimensionality reduction while preserving high flexibility, which therefore greatly improves the quality of model prediction;
	\item A residual modeling strategy accomplished through independent multi-output GP has been introduced in the posterior domain mappings in order to consider the inductive bias brought by prior domain mappings, which in turn eases model inference;
	\item Extensive comparative experiments against existing heterogeneous LMC models have been conducted on diverse heterogeneous multi-task cases to verify the superiority of our proposed model.
\end{itemize}

The remaining of this paper is organized as follows. Section~\ref{HMTGP} first defines the heterogeneous multi-task learning scenario, and thereby introduces the heterogeneous MTGP based on the LMC framework and the proposed Bayesian calibration, followed by discussing the differences to existing methods in section~\ref{sec_diff}. Thereafter, section~\ref{sec_exp} conducts comprehensive numerical experiments on two toy cases, five heterogeneous multi-task cases and the design of steam turbine exhaust to verify the superiority of the proposed model. Finally, section~\ref{sec_conclusion} provides concluding remarks regarding the study in this paper.

\section{Heterogeneous Multi-Task Gaussian Process} \label{HMTGP}
The multi-task learning over heterogeneous input domains is defined as follows. Suppose that we have a supervised training data of $T$ ($T \ge 2$) correlated tasks as $\mathcal{D}=\{\mathcal{D}^t\}_{t=1}^T=\{\mbf{X}^t, \mbf{y}^t\}_{t=1}^T$, where the inputs $\mbf{X} = \{\mbf{X}^t \in \mathbb{R}^{N^t \times D^t} \}_{t=1}^T$ and the associated outputs $\mbf{y} = \{\mbf{y}^t \in \mathbb{R}^{N^t}\}_{t=1}^T$. The \textit{heterogeneousness} in this paper means that the tasks are defined in input domains $\{\mathcal{X}^t\}_{t=1}^T$ with varying features. That is, there exists two tasks $t$ and $t'$ ($t \ne t'$) such that $\mathcal{X}^t \ne \mathcal{X}^{t'}$.\footnote{The difference to multi-view learning~\cite{sun2013survey} is that the latter usually targets on the same task from multiple views with varying dimensions (features).} For instance, the input domains might have different dimensionalities ($D^t \ne D^{t'}$), or different features (e.g., one accepts the time feature and the other is the frequency feature). Besides, the dimensions are assumed to be sorted by descending, i.e., $D^{t+1} \ge D^t$. The goal is to adopt Gaussian process to learn the mapping $\mathcal{X}^1 \times \cdots \times \mathcal{X}^T \mapsto \mathcal{Y}^1 \times \cdots \times  \mathcal{Y}^T$ for modeling these correlated tasks simultaneously and predicting the outputs $\mbf{y}_*=\{y_*^t \}_{t=1}^T$ at $T$ arbitrary test points $\mbf{x}_*=\{\mbf{x}_*^t \}_{t=1}^T$.

Besides, we further assume that a prior domain mapping (also known as nominal mapping~\cite{tao2019input}) $g_0^t(.)$ ($1 \le t \le T$) is available based on practical expert opinion. This mapping $g_0^t(.): \mathbb{R}^{D^t} \mapsto \mathbb{R}^D$ describes transformation of the $t$-th input domain $\mathcal{X}^t$ to a common domain $\mathcal{X}$ with dimensionality $D$ that satisfies $D \le D^t$. Note that in practice when we take the input domain of the task with the lowest dimension, i.e., the $T$-th task, as the common domain, the corresponding domain mapping is an identity function $g_0^T(\mbf{X}^T) = I(\mbf{X}^T) \triangleq \mbf{X}^T$.

With the above definition, we present below heterogeneous MTGPs to achieve multi-task modeling over heterogeneous input domains.

\subsection{Heterogeneous MTGP with prior domain mapping}
Since the prior domain mappings $\{g^t_0(.)\}_{t=1}^T$ are known in advance, the straightforward way is mapping these input domains to the common domain to achieve input alignment, and then performing the conventional MTGP modeling in this $D$-dimensional common domain $\mathcal{X}$. That is, for a given input $\mbf{x}^t \in \mathbb{R}^{D^t}$, we transform it through $g_0^t(.)$ to get the\textit{ aligned input} as
\begin{align}
	\bar{\mbf{x}}^t = g_0^t(\mbf{x}^t) \in \mathbb{R}^D, \quad 1 \le t \le T.
\end{align}
Consequently, the input data of all the $T$ tasks are within the same domain with dimensionality $D$.

Thereafter, we employ the LMC, a well-known MTGP model, and present the stochastic variational training framework to improve the scalability for handling massive data. Specifically, the LMC expresses each task as a linearly weighted combination of several \textit{independent} GPs, i.e.,
\begin{align}
	y^t(\bar{\mbf{x}}^t) = \sum_{q=1}^Q a_q^t f_q(\bar{\mbf{x}}^t) + \epsilon^t,
\end{align}
where $a_q^t$ is the mixing coefficient to be inferred from data; the task-specific Gaussian noise $\epsilon^t \sim \mathcal{N}(\epsilon^t|0, \nu_{\epsilon}^t)$ is independent and identically distributed (\textit{i.i.d.}); and finally, each of the $Q$ latent functions follows a GP prior
\begin{align}
	f_q(.) \sim \mathcal{GP}(m_q(.), k_q(.,.)), \quad 1 \le q \le Q,
\end{align}
with $m_q(.)$ being the mean function which often takes zero for simplicity in practice, and $k_q(.,.)$ being the kernel describing the correlations of inputs over domain $\mathcal{X}$. It is notable that the latent function $f_q(.)$ is \textit{not} task-specific and will be evaluated at all the $N = \sum_{t=1}^T N^t$ data points collected from $T$ tasks.

In order to train the LMC model, we usually take the type-II maximum likelihood strategy by maximizing the log marginal likelihood expressed as
\begin{align}
\begin{aligned}
\log p(\mbf{y}) =& \log \mathbb{E}_{p(\mbf{f})}[p(\mbf{y}|\mbf{f})] = \sum_{t=1}^T \log \mathbb{E}_{p(\mbf{f})}[ p(\mbf{y}^t|\mbf{f})],
\end{aligned}
\end{align}
where the latent function values $\mbf{f} = \{\mbf{f}_q \in \mathbb{R}^N \}_{q=1}^Q$, the $t$-th likelihood factorizes over data points as
\begin{align}
	p(\mbf{y}^t|\mbf{f}) = \prod_{i=1}^{N^t} \mathcal{N}\left(y_i^t \left| \sum_{q=1}^Q a_q^t f_q(\bar{\mbf{x}}_i^t), \nu_{\epsilon}^t \right. \right),
\end{align}
and the joint prior also factorizes over the independent latent functions as
\begin{align}
	p(\mbf{f}) = \prod_{q=1}^Q p(\mbf{f}_q) = \prod_{q=1}^Q \mathcal{N}(\mbf{f}_q|\mbf{0}, \mbf{K}_{\mbf{f}_q\mbf{f}_q}),
\end{align}
with the $N \times N$ covariance matrix $\mbf{K}_{\mbf{f}_q\mbf{f}_q} = k_q(\mbf{X}, \mbf{X})$, the determinant and inversion of which however become time-consuming due to the cubic space complexity $\mathcal{O}(N^3)$ when handling massive data from multiple tasks.

Hence, in order to improve the scalability of LMC, we introduce the sparse approximation that adopts $M_q$ ($M_q \ll N$) inducing variables $\mbf{u}_q$ at the pseudo inputs $\mbf{Z}_q \in \mathbb{R}^{M_q \times D}$ to be the sufficient statistics of $\mbf{f}_q$. As a result, we arrive at the following conditional
\begin{align}
	p(\mbf{f}_q|\mbf{u}_q) = \mathcal{N}(\mbf{f}_q|\mbf{K}_{\mbf{f}_q\mbf{u}_q} \mbf{K}_{\mbf{u}_q\mbf{u}_q}^{-1} \mbf{u}_q, \mbf{K}_{\mbf{f}_q\mbf{f}_q} - \mbf{K}_{\mbf{f}_q\mbf{u}_q} \mbf{K}_{\mbf{u}_q\mbf{u}_q}^{-1} \mbf{K}_{\mbf{f}_q\mbf{u}_q}^{\mathsf{T}}),
\end{align}
which has only $\mathcal{O}(NM_q^2)$ complexity with the covariances $\mbf{K}_{\mbf{f}_q\mbf{u}_q} = k_q(\mbf{X}, \mbf{Z}_q) \in \mathbb{R}^{N \times M_q}$ and $\mbf{K}_{\mbf{u}_q\mbf{u}_q} = k_q(\mbf{Z}_q, \mbf{Z}_q) \in \mathbb{R}^{M_q \times M_q}$. 

The interested non-Gaussian posterior $p(\mbf{f}, \mbf{u}|\mbf{y}) = \prod_{q=1}^Q p(\mbf{f}_q| \mbf{u}_q) p(\mbf{u}_q|\mbf{y})$ in sparse GP however is intractable. Hence, to obtain both the marginal likelihood and the posterior, we resort to variational inference that adopts a Gaussian variational posterior $q(\mbf{u}_q) = \mathcal{N}(\mbf{u}_q|\mbf{m}_q, \mbf{S}_q)$ as an approximation to minimize the Kullback-Leibler divergence $\mathrm{KL}[q(\mbf{f}, \mbf{u}) || p(\mbf{f}, \mbf{u}| \mbf{y})]$. Consequently, it is equivalent to maximizing the evidence lower bound (ELBO) of $\log p(\mbf{y})$ as
\begin{align} \label{eq_elbo_hsvlmc_g0}
	\begin{aligned}
		\mathcal{L} =& \mathbb{E}_{q(\mbf{f})} [\log p(\mbf{y}|\mbf{f})] - \mathrm{KL}[q(\mbf{u}) || p(\mbf{u})] \\
		=& \sum_{t=1}^T \left[\sum_{i=1}^{N^t} \log \mathcal{N}\left(y_i^t \left|\sum_{q=1}^Q a_q^t \mu_{q,i}^t, \nu_{\epsilon}^t \right. \right) \right. \\
		& \left. - \frac{1}{2 \nu_{\epsilon}} \sum_{q=1}^Q (a_q^t)^2 \nu_{q,i}^t \right] - \sum_{q=1}^Q \mathrm{KL}[q(\mbf{u}_q) || p(\mbf{u}_q)],
	\end{aligned}
\end{align}
where the Gaussian posterior $q(\mbf{f})$ factorizes as $q(\mbf{f}) = \prod_{q=1}^Q q(\mbf{f}_q) = \prod_{q=1}^Q \int p(\mbf{f}_q|\mbf{u}_q) q(\mbf{u}_q) d\mbf{u}_q$, the mean $\mu_{q,i}^t$ and variance $\nu_{q,i}^t$ come from the variational Gaussian $q(f_q(\bar{\mbf{x}}_i^t)) = \int p(f_q(\bar{\mbf{x}}_i^t)|\mbf{u}_q) q(\mbf{u}_q) d\mbf{u}_q$, and are respectively expressed as
\begin{align}
	\mu_{q,i}^t =& k_q(\bar{\mbf{x}}_i^t, \mbf{Z}_q) \mbf{K}_{\mbf{u}_q\mbf{u}_q}^{-1} \mbf{m}_q, \label{eq_mu_qi^t} \\
	\nu_{q,i}^t =& k_q(\bar{\mbf{x}}_i^t, \bar{\mbf{x}}_i^t) + k_q(\bar{\mbf{x}}_i^t, \mbf{Z}_q) \mbf{K}_{\mbf{u}_q\mbf{u}_q}^{-1} [\mbf{S}_q \mbf{K}_{\mbf{u}_q\mbf{u}_q}^{-1} - \mbf{I}] k_q^{\mathsf{T}}(\bar{\mbf{x}}_i^t, \mbf{Z}_q). \label{eq_nu_qi^t}
\end{align}

Furthermore, since the first expectation term in the right-hand side of $\mathcal{L}$~\eqref{eq_elbo_hsvlmc_g0} factorizes over data points, the ELBO could have an unbiased estimation on a subset $\mathcal{B}^t$ of the training data $\mathcal{D}^t$ with $|\mathcal{B}^t| \ll N^t$ as
\begin{align} \label{eq_elbo_hsvlmc_g0_B}
	\begin{aligned}
		\mathcal{L} \approx& \sum_{t=1}^T \frac{N^t}{|\mathcal{B}^t|} \left[\sum_{i\in \mathcal{B}^t} \log \mathcal{N}\left(y_i^t \left|\sum_{q=1}^Q a_q^t \mu_{q,i}^t, \nu_{\epsilon}^t \right. \right) \right. \\
		&\left. - \frac{1}{2 \nu_{\epsilon}} \sum_{q=1}^Q (a_q^t)^2 \nu_{q,i}^t \right] - \sum_{q=1}^Q \mathrm{KL}[q(\mbf{u}_q) || p(\mbf{u}_q)],
	\end{aligned}
\end{align}
which therefore further reduces the model complexity to $\mathcal{O}(|\mathcal{B}|M_q^2)$ with $|\mathcal{B}| = \sum_{t=1}^T |\mathcal{B}^t|$, and could be efficiently tuned through stochastic optimizer, e.g., Adam~\cite{kingma2015adam}.

Besides, given the normal prior $p(\mbf{u}_q) = \mathcal{N}(\mbf{u}_q|\mbf{0}, \mbf{K}_{\mbf{u}_q\mbf{u}_q})$, the KL terms in the right-hand side of~\eqref{eq_elbo_hsvlmc_g0} have closed-form expressions. For example, for $\mathrm{KL}[q(\mbf{u}_q)||p(\mbf{u}_q)]$, we have
\begin{align} \label{eq_kl_u_q}
	\begin{aligned}
		\mathrm{KL}[q(\mbf{u}_q)||p(\mbf{u}_q)] =& \frac{1}{2} \left[\log \frac{|\mbf{K}_{\mbf{u}_q\mbf{u}_q}|}{|\mbf{S}_q|} + \mbf{m}_q^{\mathsf{T}} \mbf{K}_{\mbf{u}_q\mbf{u}_q}^{-1} \mbf{m}_q \right. \\
		&\left. + \mathrm{Tr} [\mbf{K}_{\mbf{u}_q\mbf{u}_q}^{-1} \mbf{S}_q - \mbf{I}] \right].
	\end{aligned}
\end{align}
Combining~\eqref{eq_elbo_hsvlmc_g0_B} and~\eqref{eq_kl_u_q}, we finally obtain the analytical ELBO for the heterogeneous stochastic variational LMC via prior domain mapping, which is denoted as \texttt{HSVLMC-g0}.

Finally, for the prediction of the $t$-th task at a test point $\mbf{x}_*^t$, we have
\begin{align}
	\begin{aligned}
		p(y_*^t|\mathcal{D}, \mbf{x}_*^t) =& \int p(y_*^t|\mbf{f}_*) p(\mbf{f}_*|\mbf{u}) q(\mbf{u}) d\mbf{f}_* d\mbf{u} \\
		=& \mathcal{N}(y_*^t|\mu_*^t, \nu_*^t),
	\end{aligned}
\end{align}
where $\mbf{f}_* = \{f_{q,*} \}_{q=1}^Q$ collects the $Q$ latent function values at point $\mbf{x}_*^t$. Note that the prediction mean $\mu_*^t = \sum_{q=1}^Q a_q^t \mu_{q,*}^t$ and the variance $\nu_*^t = \sum_{q=1}^Q (a_q^t)^2 \nu_{q,*}^t + \nu_{\epsilon}^t$ with $\mu_{q,*}^t$ and $\nu_{q,*}^t$ taking the forms in~\eqref{eq_mu_qi^t} and~\eqref{eq_nu_qi^t} with $\bar{\mbf{x}}_i^t$ replaced by $\bar{\mbf{x}}_*^t = g_0^t(\mbf{x}_*^t)$.

\subsection{Heterogeneous MTGP with Bayesian calibration}
Directly projecting the heterogeneous input domains into a low-dimensional common domain through prior domain mappings without calibration will induce information loss. For instance, a dataset can be linearly classified in the original high-dimensional input space; but it may become a nonlinear classification problem in the low-dimensional manifold due to information embedding. Hence, we here propose the \texttt{HSVLMC} model with Bayesian calibration to learn preferred domain mappings from data for (i) the alignment of input domains and (ii) the model calibration to alleviate information loss as well as enabling model flexibility during domain alignment.

To this end, we are interested in inferring informative posterior of the aligned inputs $\{\bar{\mbf{x}}_i^t \}_{t=1}^{N^t}$ to accomplish model calibration. Given the training data, we adopt variational inference to have variational Gaussians for the aligned inputs of the $t$-th task as
\begin{align} \label{eq_qx}
	q(\bar{\mbf{X}}^t) = \prod_{i=1}^{N^t} q(\bar{\mbf{x}}_i^t) = \prod_{i=1}^{N^t} \mathcal{N}(\bar{\mbf{x}}_i^t|\bm{\mu}_{g,i}^t, \mathrm{diag}[\bm{\nu}_{g,i}^t]),
\end{align}
where the aligned input $\bar{\mbf{x}}_i^t = g^t(\mbf{x}_i^t)$ now is mapped through the \textit{posterior} domain mapping $g^t(.)$. Note that we here assume that the Gaussians in~\eqref{eq_qx} factorize over both data points and dimensions for simplicity. Similarly, the joint posterior now is approximated as
\begin{align}
	\begin{aligned}
		p(\mbf{f}, \mbf{u}, \bar{\mbf{X}} | \mbf{y}) \approx & q(\mbf{f}, \mbf{u}, \bar{\mbf{X}}) = p(\mbf{f} | \mbf{u}, \bar{\mbf{X}}) q(\mbf{u}) q(\bar{\mbf{X}}) \\
		=& \prod_{q=1}^Q p(\mbf{f}_q | \mbf{u}_q, \bar{\mbf{X}}) q(\mbf{u}_q) \prod_{t=1}^T q(\bar{\mbf{X}}^t).
	\end{aligned}
\end{align}
Thereafter, similar to~\eqref{eq_elbo_hsvlmc_g0}, by reformulating the KL divergence $\mathrm{KL}[q(\mbf{f}, \mbf{u}, \bar{\mbf{X}})||p(\mbf{f}, \mbf{u}, \bar{\mbf{X}} | \mbf{y}) ]$, we arrive at the ELBO as
\begin{align} \label{eq_elbo_hsvlmc}
	\begin{aligned}
		\mathcal{L} =& \mathbb{E}_{q(\mbf{f}|\bar{\mbf{X}}) q(\bar{\mbf{X}})} [\log p(\mbf{y}|\mbf{f})] - \mathrm{KL}[q(\mbf{u}) || p(\mbf{u})] \\
		&- \mathrm{KL}[q(\bar{\mbf{X}}) || p(\bar{\mbf{X}})],
	\end{aligned}
\end{align}
where the posterior $q(\mbf{f}|\bar{\mbf{X}}) = \int p(\mbf{f}, \mbf{u}|\bar{\mbf{X}}) q(\mbf{u}) d\mbf{u} $, and the KL term factorizes over tasks as $\mathrm{KL}[q(\bar{\mbf{X}}) || p(\bar{\mbf{X}})] = \sum_{t=1}^T \mathrm{KL}[q(\bar{\mbf{X}}^t) || p(\bar{\mbf{X}}^t)]$. The ELBO $\mathcal{L}$ in~\eqref{eq_elbo_hsvlmc} usually has no analytical expression due to the stochastic inputs $\bar{\mbf{X}}$. Hence, we could take the reparameterization trick~\cite{kingma2013auto} to estimate the ELBO by sampling from Gaussians.

In the above ELBO, in contrast to the posterior $q(\bar{\mbf{X}}^t)$ in~\eqref{eq_qx}, the prior of aligned inputs is assumed to follow the following Gaussian distribution factorized as
\begin{align} \label{eq_px}
	p(\bar{\mbf{X}}^t) = \prod_{i=1}^{N^t} p(\bar{\mbf{x}}_i^t) = \prod_{i=1}^{N^t} \mathcal{N}(\bar{\mbf{x}}_i^t|\bm{\mu}_{g_0, i}^t \triangleq g_0^t(\mbf{x}_i^t), \nu_{g_0}^t \mbf{I}).
\end{align}
For this prior distribution, we include the prior domain mapping $g_0^t(.)$ into the mean to consider the inductive bias. Besides, we particularly introduce a learnable variance $\nu_{g_0}^t$ to improve model flexibility. It is observed that a small $\nu_{g_0}^t$ raises nearly deterministic domain transformation, and consequently, the KL penalty in~\eqref{eq_elbo_hsvlmc} pushes the posterior mapping $g^t(.)$ towards the prior mapping $g^t_0(.)$; contrarily, a large variance $\nu_{g_0}^t$ allows flexible domain transformation to account for powerful model calibration. Note that when we are taking the input domain of $T$-th task with the lowest dimension as the common domain, i.e., $\mathcal{X} \triangleq \mathcal{X}^T$, then the posterior $q(\bar{\mbf{X}}^T)$ and the prior $p(\bar{\mbf{X}}^T)$ vanish due to the identity mapping.

Besides, for the unknown mean $\bm{\mu}_{g,i}^t$ and variance $\bm{\nu}_{g,i}^t$ in the posterior $q(\bar{\mbf{x}}_i^t)$ in~\eqref{eq_qx}, we could treat them as hyperparameters and infer from data. This \textit{data-dependent} parameterization however raises $2\times ND$ hyperparameters in total and is only available at training phase. 

Alternatively, the posterior $q(\bar{\mbf{x}}_i^t)$ can be derived and accomplished through the complete GP paradigm. That is, we come up with the GP-inspired stochastic domain mapping $g^t(.)$ by introducing an additional independent multi-output sparse GP (MSGP) to learn the mapping from $\chi^t \in \mathbb{R}^{D^t}$ to $\chi \in \mathbb{R}^{D}$. Consequently, we obtain
\begin{align}
	[\bm{\mu}_{g,i}^t, \bm{\nu}_{g,i}^t] =& \mathrm{MSGP}^t(\mbf{x}_i^t, \mbf{u}_g^t), \label{eq_msgp}\\
	\bm{\mu}_{g,i}^t =& \bm{\mu}_{g,i}^t  + g_0^t(\mbf{x}_i^t), \label{eq_qx_mu_gp_residual}
\end{align}
where $\mathrm{MSGP}^t(.): \mathbb{R}^{D^t} \mapsto \mathbb{R}^D$ models the $D$ outputs independently for the $t$-th input domain $\mathcal{X}^t$, and it simply takes the zero-mean GP $\mathcal{GP}(0, k^t(.,.))$ shared across $D$ outputs; and $\mbf{u}_g^t = \{\mbf{u}_g^{t,d} \in \mathbb{R}^{M_g^t}\}_{d=1}^D$ contains the inducing variables at the shared pseudo inputs $\mbf{Z}^t_g \in \mathbb{R}^{M_g^t \times D^t}$ for the overall $D$ outputs.\footnote{The number and location of pseudo inputs can vary for $D$ outputs.} Besides, note that we particularly adopt the \textit{residual} formulation in~\eqref{eq_qx_mu_gp_residual} to express the posterior mean, which therefore introduces the prior inductive bias to ease model training. Finally, similar to~\eqref{eq_mu_qi^t} and~\eqref{eq_nu_qi^t}, given the variational Gaussian posterior $q(\mbf{u}_g^{t,d}) = \mathcal{N}(\mbf{u}^{t,d}_g|\mbf{m}_g^{t,d}, \mbf{S}_g^{t,d})$, we have
\begin{align}
	\mu_{g,i}^{t,d} =& k^t(\mbf{x}_i^t, \mbf{Z}_g^t) \mbf{K}_{\mbf{u}^t\mbf{u}^t}^{-1} \mbf{m}_g^{t,d} + \mu_{g_0, i}^{t,d}, \\
	\nu_{g,i}^{t,d} =& k^t(\mbf{x}_i^t, \mbf{x}_i^t) + k^t(\mbf{x}_i^t, \mbf{Z}_g^t) \mbf{K}_{\mbf{u}^t\mbf{u}^t}^{-1} [\mbf{S}_g^{t,d} \mbf{K}_{\mbf{u}^t\mbf{u}^t}^{-1} - \mbf{I}] k^{t}(\mbf{Z}_g^t, \mbf{x}_i^t),
\end{align}
where $\mu_{g_0, i}^{t,d}$ is the $d$-dimensional value of $g_0^t(\mbf{x}_i^t)$, and the covariance $\mbf{K}_{\mbf{u}^t\mbf{u}^t} = k^t(\mbf{Z}_g^t, \mbf{Z}_g^t) \in \mathbb{R}^{M_g^t \times M_g^t}$. 

When the posterior $q(\bar{\mbf{X}}^t)$ takes the MSGP parameterization, the ELBO in~\eqref{eq_elbo_hsvlmc} has an additional KL term as
\begin{align} \label{eq_elbo_hsvlmc_final}
	\begin{aligned}
		\mathcal{L} =& \mathbb{E}_{q(\mbf{f}|\bar{\mbf{X}}) q(\bar{\mbf{X}})} [\log p(\mbf{y}|\mbf{f})] - \mathrm{KL}[q(\mbf{u}) || p(\mbf{u})] \\
		&- \mathrm{KL}[q(\mbf{u}_g) || p(\mbf{u}_g)] - \mathrm{KL}[q(\bar{\mbf{X}}) || p(\bar{\mbf{X}})],
	\end{aligned}
\end{align}
where the set $\mbf{u}_g = \{\mbf{u}_g^t \}_{t=1}^T$ contains the inducing variables for the $T$ GP-inspired domain mappings; the KL term regarding $\mbf{u}_g$ factorizes as $\mathrm{KL}[q(\mbf{u}_g) || p(\mbf{u}_g)] = \sum_{t=1}^T \sum_{d=1}^D \mathrm{KL}[q(\mbf{u}_g^{t,d}) || p(\mbf{u}_g^{t,d})]$, with the component analytically calculated as
\begin{align} \label{eq_kl_u_g^td}
	\begin{aligned}
		&\mathrm{KL}[q(\mbf{u}_q^{t,d})||p(\mbf{u}_q^{t,d})] \\
		=& \frac{1}{2} \left[\log \frac{|\mbf{K}_{\mbf{u}_q^{t,d}\mbf{u}_q^{t,d}}|}{|\mbf{S}_q^{t,d}|} + (\mbf{m}_q^{t,d})^{\mathsf{T}} \mbf{K}_{\mbf{u}_q^{t,d}\mbf{u}_q^{t,d}}^{-1} \mbf{m}_q^{t,d} \right. \\
		&\left. + \mathrm{Tr} [\mbf{K}_{\mbf{u}_q^{t,d}\mbf{u}_q^{t,d}}^{-1} \mbf{S}_q^{t,d} - \mbf{I}] \right].
	\end{aligned}
\end{align}
and the KL term regarding $\bar{\mbf{X}}$ factorizes as $\mathrm{KL}[q(\bar{\mbf{X}}) || p(\bar{\mbf{X}})] = \sum_{t=1}^T \sum_{i=1}^{N^t} \mathrm{KL}[q(\bar{\mbf{x}}_i^t) || p(\bar{\mbf{x}}_i^t)]$, with the component analytically calculated as
\begin{align} \label{eq_kl_x_i^t}
	\begin{aligned}
		&\mathrm{KL}[q(\bar{\mbf{x}}_i^t) || p(\bar{\mbf{x}}_i^t)] \\
		=& \frac{1}{2} \left(D \log \nu_{g_0}^t - \sum_{d=1}^D \log \nu_{g,i}^{t,d} - D \right. \\
		&\left. + \frac{1}{\nu_{g_0}^t} \sum_{d=1}^D \nu_{g,i}^{t,d} + \frac{1}{\nu_{g_0}^t} \sum_{d=1}^D (\mu_{g,i}^{t,d} - \mu_{g_0,i}^{t,d})^2 \right).
	\end{aligned}
\end{align}
Note that akin to~\eqref{eq_elbo_hsvlmc_g0_B}, the ELBO in~\eqref{eq_elbo_hsvlmc_final} can be estimated efficiently on subsets $\{\mathcal{B}^t \}_{t=1}^T$ randomly sampled from $\{\mathcal{D}^t \}_{t=1}^T$.

Finally, we again highlight that the main difference of the proposed \texttt{HSVLMC} to the simple \texttt{HSVLMC-g0} is the additional Bayesian calibration conducted through GP-inspired posterior domain mappings. This Bayesian calibration takes into account the effect of dimensionality reduction, and thus brings benefits and flexibility for model enhancement, which will be verified in the numerical experiments.

\section{Differences to other calibrations} \label{sec_diff}
Except for the Bayesian calibration proposed in this paper, there are also some other calibrations in literature to tackle multi-task/multi-fidelity modeling with heterogeneous inputs. Inspired by the idea of space mapping~\cite{bandler2004space, robinson2008surrogate, rayas2016power}, the input mapping calibration (IMC)~\cite{tao2019input} attempts to find a better linear mapping than the prior linear domain mapping. Specifically, given a high-fidelity task with $N^h$ data $\mathcal{D}^h = \{\mbf{X}^h, \mbf{y}^h \}$ and a related low-fidelity task with $N^l$ data $\mathcal{D}^l = \{\mbf{X}^l, \mbf{y}^l \}$, the IMC approach takes a linear transformation
\begin{align}
	g^h(\mbf{x}^h) = \mbf{A} \mbf{x}^h + \mbf{b}
\end{align}
to project the high-fidelity inputs into the low-dimensional low-fidelity input space. In order to learn the transformation matrix $\mbf{A} \in \mathbb{R}^{D^l \times D^h}$ and the bias correction $\mbf{b} \in \mathbb{R}^{D^l}$ from data, the IMC minimizes the discrepancy of task outputs at the high fidelity data points as
\begin{align}
	\begin{aligned}
		(\mbf{A}_{\mathrm{opt}}, \mbf{b}_{\mathrm{opt}}) =& \arg \min_{\mbf{A}, \mbf{b}} \sum_{i=1}^{N^h} (y_i^h - y^l(g^h(\mbf{x}^h)))^2 \\
		&+ \alpha R([\mbf{A};\mbf{b}], [\mbf{A}_0;\mbf{b}_0]),
	\end{aligned}
\end{align}
where $\mbf{A}_0$ and $\mbf{b}_0$ are the known parameters of the prior linear domain mapping $g_0^h(.)$; $\alpha$ is the penalty factor and $R(.,.)$ is the regularizer; and finally, $y^l(.)$ is the low-fidelity function which is assumed to be known in advance. The IMC approach has limitations in handling complicated scenarios, since (i) it only considers the simple linear transformation of input space; (ii) the input calibration is independent of the subsequent model training, i.e., it is not an end-to-end model; and (iii) the low fidelity function $y^l(.)$ is usually unknown in practice. Since the prior domain mappings are known in advance, the \texttt{HSVLMC-g0} looks like a more reasonable direct implementation in comparison to the IMC method.

Similar to \texttt{HSVLMC-g0}, Li et al.~\cite{li2016integrating} proposed directly incorporating the prior domain mapping into the multi-fidelity modeling. Recently, a multi-fidelity GP has been proposed by using a different calibration approach~\cite{hebbal2021multi}. Specifically, the presented embedded mapping employs a GP to directly model the prior domain mapping
\begin{align} \label{eq_prior_map_gp}
	g^t_0(.) \sim \mathcal{GP}(m^t_0(.), k^t_0(.,.)).
\end{align}
That is, the mapped inputs $\bar{\mbf{X}} = \{\bar{\mbf{X}}^t = g_0^t(\mbf{X}^t) \}_{t=1}^T$ become the observed outputs, which could be modeled by independent multi-output GP. Hence, for the more general multi-task scenario with $T$ correlated tasks, we derive the following ELBO in the framework of stochastic variational LMC as
\begin{align} \label{eq_elbo_hsvlmc_em}
	\begin{aligned}
		\mathcal{L} =& \mathbb{E}_{q(\mbf{f})} [\log p(\mbf{y}|\mbf{f})] + \mathbb{E}_{q(\bar{\mbf{F}})} [\log p(\bar{\mbf{X}}|\bar{\mbf{F}})] \\
		&- \mathrm{KL}[q(\mbf{u}) || p(\mbf{u})] - \mathrm{KL}[q(\bar{\mbf{U}}) || p(\bar{\mbf{U}})],
	\end{aligned}
\end{align}
where $\bar{\mbf{F}} = \{\bar{\mbf{F}}^t = g_0^t(\mbf{X}^t)\in \mathbb{R}^{N^t \times D} \}_{t=1}^T$ with each element representing the latent function values at training points $\mbf{X}^t$ for $D$ outputs; $\bar{\mbf{U}} = \{\bar{\mbf{U}}^t \in \mathbb{R}^{\bar{M}^{t} \times D} \}_{t=1}^T$ represents the related inducing variables of each task for $D$ outputs; and akin to $q(\mbf{f})$, the posterior $q(\bar{\mbf{F}}) = \int p(\bar{\mbf{F}}|\bar{\mbf{U}}) q(\bar{\mbf{U}}) d\bar{\mbf{U}}$. This model using embedded mapping is denoted as \texttt{HSVLMC-EM}. In comparison to the IMC method, (i) this \texttt{HSVLMC-EM} is capable of providing flexible input transformation rather than the simple linear transformation;\footnote{The \texttt{HSVLMC-EM} can recover the IMC idea by using a linear kernel $k_0^t(.,.)$ in~\eqref{eq_prior_map_gp} to achieve the similar linear embedding.} and (ii) it is an end-to-end modeling framework. Besides, in comparison to the \texttt{HSVLMC-g0}, the \texttt{HSVLMC-EM} takes into account the uncertainty of prior domain mappings. But directly modeling the prior domain mapping would weaken the role of calibration, which will be demonstrated in the following numerical experiments.

\section{Numerical experiments}
\label{sec_exp}

This section first investigates the methodological characteristics of the proposed \texttt{HSVLMC} model on two toy cases. Thereby,  we perform a comprehensive comparison study against existing competitors on five real-world heterogeneous multi-task/-fidelity cases with different characteristics, followed by discussions regarding the impact of training size, dimensionality and task correlation. 

The models are implemented within the GPflow package~\cite{matthews2017gpflow} using Tensorflow, and the above numerical experiments are conducted on a Linux workstation with TITAN RTX GPU. The detailed configurations for these numerical experiments are provided in Appendix~\ref{app_confg}. For assessing the quality of predictive distribution, we employ two loss criteria in this paper. The first is standardized mean square error (SMSE). For the $t$-th task, given $N^{t}_*$ test points $\{\mbf{x}_{*,i}^t, y_{*,i}^t\}_{i=1}^{N^{t}_*}$, the SMSE criterion is defined as
\begin{align}
	\mathrm{SMSE} = \frac{\sum_{i=1}^{N^{t}_*}(y_{*,i}^t - \mu_{*,i}^t)^2}{N^{t}_* \times y_{\nu}^t},
\end{align}
where $\mu_{*,i}^t$ is the prediction mean at test point $\mbf{x}_{*,i}^t$, and $y_{\nu}^t = \mathrm{var}(\mbf{y}^t)$ represents the variance of training observations for the $t$-th task. It is found that the SMSE indicates the difference between the prediction mean and the true observation, and particularly, it equals to one when the GP simply uses the mean of $\mbf{y}^t$, i.e., $y_{\mu}^t$, as prediction. The second loss criterion is standardized mean log loss (SMLL) defined as
\begin{align}
	\begin{aligned}
		\mathrm{SMLL} =& \frac{1}{N^{t}_*} \sum_{i=1}^{N^{t}_*} [\log \mathcal{N}(y_{*,i}^t|y_{\mu}^t, y_{\nu}^t) - \log \mathcal{N}(y_{*,i}^t|\mu_{*,i}^t, \nu_{*i}^t)],
	\end{aligned}
\end{align}
where $\mu_{*,i}^t$ and $\nu_{*,i}^t$ are the prediction mean and variance at test point $\mbf{x}_{*,i}^t$,\footnote{It is notable that when the predictive distribution is non-Gaussian, we simply estimate $\mu_{*,i}^t$ and $\nu_{*,i}^t$ from samples.} and the log likelihood is calculated as
\begin{align*}
	\log \mathcal{N}(y_{*,i}^t|\mu_{*,i}^t, \nu_{*,i}^t) = -\frac{1}{2} \left[\log(2\pi \nu_{*,i}^t) + \frac{(y_{*,i}^t - \mu_{*,i}^t)^2}{\nu_{*,i}^t} \right].
\end{align*}
Different from the SMSE criterion, the SMLL criterion could quantify the quality of predictive distribution, and it is usually negative for good probabilistic predictions; particularly, it equals to zero when the GP adopts the mean and variance of training data $\mbf{y}^t$ as predictions. For the employed SMSE and SMLL criteria, lower is better.

Finally, note that the cases studied in this section has the following characteristics: (i) each of them has two tasks (outputs) $y^1$ and $y^2$; (ii) the input domain of the low-dimensional task $y^2$ is chose as the common domain $\mathcal{X}$, i.e., $g_0^2(.) = I(.)$; (iii) all the multi-task cases except $\mathtt{sarcos^b}$ follow the \textit{asymmetric} scenario defined in~\cite{liu2018remarks}, i.e., the task $y^1$ with higher dimensions only has a few data points, and we attempt to transfer knowledge from the low-dimensional task $y^2$ with abundant training data in order to improve the prediction for $y^1$;\footnote{This is often encountered in multi-fidelity modeling or transfer learning.} and finally, (iv) the $\mathtt{sarcos^b}$ case follows the \textit{symmetric} scenario defined in~\cite{liu2018remarks}, i.e., the two tasks are of equal importance and we attempt to improve their predictions simultaneously. Therefore, we report the SMSE and SMLL results averaged over tasks on the $\mathtt{sarcos^b}$ case; but we only report the SMSE and SMLL results for the target task $y^1$ on the remaining cases.

\subsection{Toy cases}
\textbf{The noisy case.} This case is adopted and warped from~\cite{liu2021scalable}, with two tasks generated from four latent functions as
\begin{align}
	\begin{aligned}
		y^1(\mbf{x}) =& 0.5f_1(x_1) - 0.4f_2(x_1) + 0.6f_3(x_1) \\
		&+ 0.6f_4(x_1) + \epsilon(x_2), \\
		y^2(x_1) =& -0.3f_1(x_1) + 0.43f_2(x_1) - 0.5f_3(x_1) \\
		&+ 0.1f_4(x_1),
	\end{aligned}
\end{align}
where the first input $x_1 \in [-5, 5]$, and the four latent functions to generate the two tasks are expressed respectively as
\begin{align}
	\begin{aligned}
		f_1(x) =& 0.5\sin(3x) + x, \, f_2(x) = 3\cos(x) - x, \\
		f_3(x) =& 2.5\cos(5x-1), \, f_4(x) = \sin(1.5x),
	\end{aligned}
\end{align}
and the noise for task $y^1$ is $\epsilon(x_2) \sim \mathcal{N}(\epsilon|0, 0.04)$. Note that we add the additional virtual input $x_2$ in the noise $\epsilon(.)$ for $y^1$ to construct the heterogeneous input domains with the domain mappings $g_0^1(x_1, x_2) = x_1$ and $g_0^2(x_1) = I(x_1)$. 

\begin{figure}[t!]
	\centering
	\includegraphics[width=0.4\textwidth]{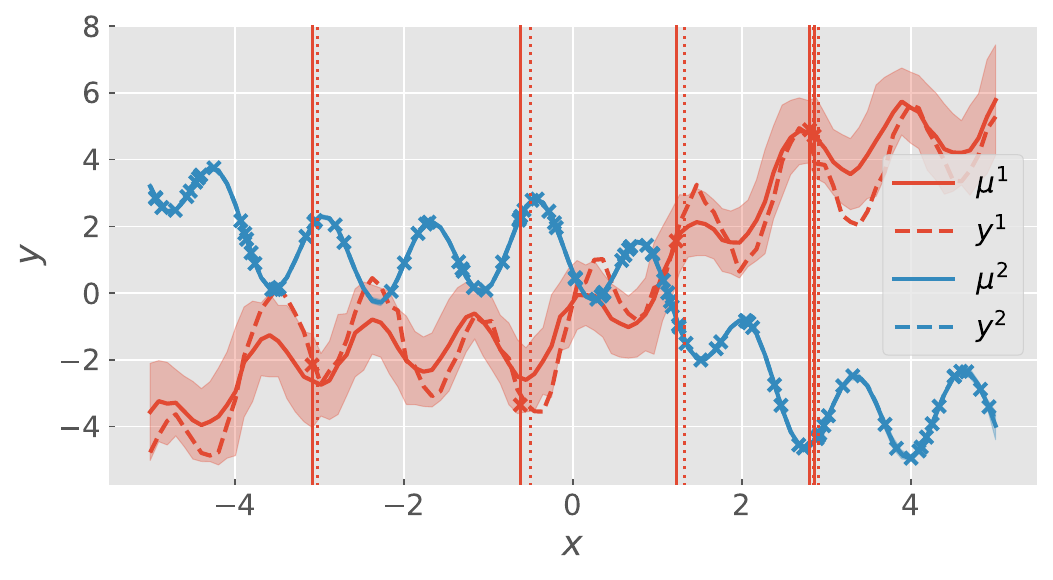}
	\caption{Illustration of the predictions of \texttt{HSVLMC} on the noisy case, with the shaded region representing 95\% confidence interval. Note that the crosses are training data, the dot vertical lines indicate the mean $\mu_g^1$ of aligned inputs via the GP-inspired domain mapping $g^1(.)$, while the solid vertical lines represent the mean $\mu_{g_0}^1$ of aligned inputs via the prior domain mapping $g_0^1(.)$.}
	\label{fig_case1}
\end{figure}

For this case, we randomly generate $N^1=5$ points for the first two-dimensional task $y^1$ and $N^2=100$ points for the second one-dimensional task $y^2$, and have separate 100 test points for each task. The goal of this case is to verify the ability of proposed \texttt{HSVLMC} to eliminate the interference of useless input $x_2$ for achieving knowledge transfer between the two tasks with heterogeneous input domains. Fig.~\ref{fig_case1} illustrates the predictions of \texttt{HSVLMC} for the two tasks on this case. Note that since the virtual input $x_2$ is useless for $y^1$, we therefore could depict the two tasks in the same figure. 

It is observed that the proposed model well fits the first task $y^1$ and reasonably quantifies the uncertainty using only 5 training points, by transferring knowledge from the second one-dimensional task $y^2$. Besides, it is known that the additional input $x_2$ has no contribution to the output $y^1$, indicating that the prior domain mapping $g_0^1: \mbf{x} \mapsto x_1$ induces no information loss. Consequently, the proposed \texttt{HSVLMC} learns a tiny variance $\nu_{g_0}^1=1.5\times 10^{-3}$ from data in order to further push the posterior $q(\bar{\mbf{X}})$ to approximate the prior $p(\bar{\mbf{X}})$. This is indicated in Fig.~\ref{fig_case1}: the locations of aligned input $\mbf{x}$ (dot vertical lines) are close to the locations of input $\mbf{x}$ aligned through $g_0^1$ (solid vertical lines).

\textbf{The multi-fidelity case.} This case describes the Park multi-fidelity problem~\cite{xiong2013sequential, hebbal2021multi}, with the modifications being that the high-fidelity and low-fidelity simulations are conducted in heterogeneous input domains. To this end, some relatively unimportant inputs are ignored in the low-fidelity function. The four-dimensional high-fidelity task defined in the unit input space is expressed as
\begin{align}
	\begin{aligned}
		y^1(x_1,x_2,x_3,x_4) =& \frac{x_1}{2}\left(\sqrt{1+(x_2+x_3^2)\frac{x_4}{x_1^2}}-1\right) \\
		&+ (x_1+3x_4)\exp(1+\sin(x_3)).
	\end{aligned}
\end{align}
The two-dimensional low-fidelity task defined in the unit space ignores the inputs $x_1$ and $x_2$,\footnote{Sobol sensitivity analysis~\cite{al2019meta} on the task $y^1$ reveals that the two inputs $x_3$ and $x_4$ contribute more to the output.} and is expressed as
\begin{align}
	\begin{aligned}
		y^2(x_3,x_4) =& \left(1+\frac{\sin(x_3)}{10}\right) y^1(0.5,0.5,x_3,x_4) \\
		&- 2x_3 + x_4^2 + 0.75.
	\end{aligned}
\end{align}
The related prior domain mappings are $g_0^1(x_1,x_2,x_3,x_4) = [x_3, x_4]^{\mathsf{T}}$ and $g_0^2(.) = I(.)$.

\begin{figure*}[t!]
	\centering
	\includegraphics[width=0.7\textwidth]{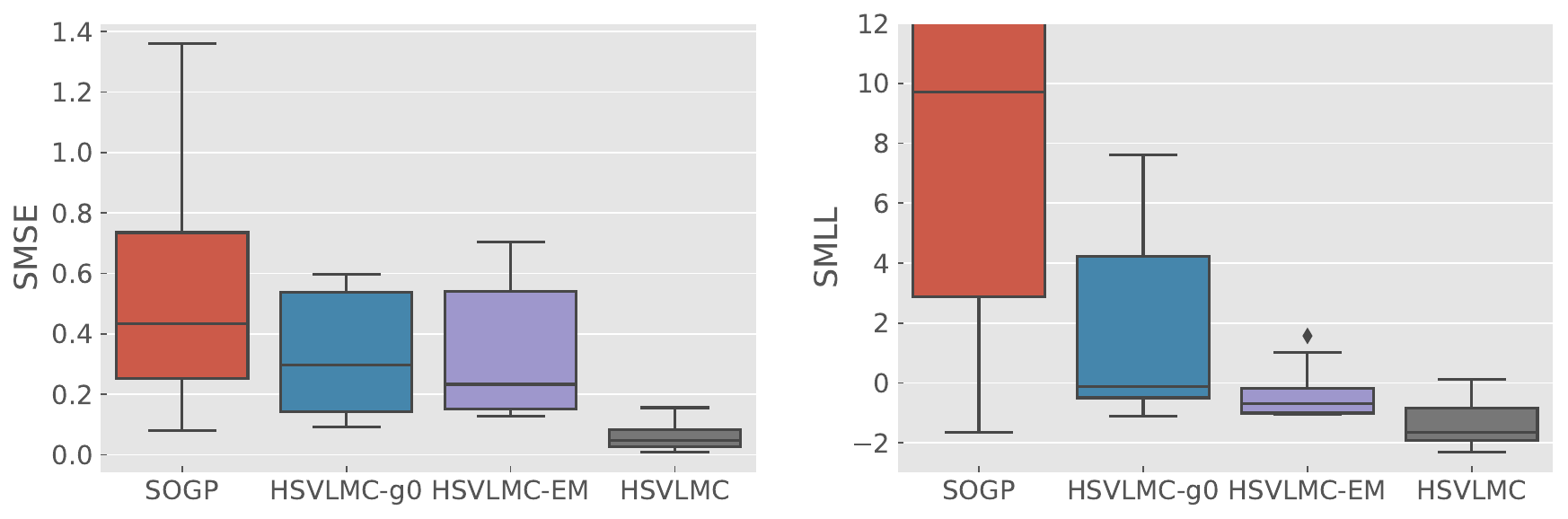}
	\caption{Boxplots of different GP models on the multi-fidelity case in terms of the SMSE and SMLL criteria.}
	\label{fig_case2_boxplot}
\end{figure*}

We generate $N^1=6$ random training points for the high-fidelity task $y^1$ and $N^2=100$ training points for the low-fidelity task $y^2$, and compare the proposed \texttt{HSVLMC} against the competitors including (i) the single output GP (\texttt{SOGP}); (ii) the \texttt{HSVLMC-g0} that directly uses the prior domain mappings like~\cite{li2016integrating}; and (iii) the \texttt{HSVLMC-EM} using the embedded mapping strategy deeloped in~\cite{hebbal2021multi}. Fig.~\ref{fig_case2_boxplot} depicts the boxplots of different GP models on this multi-fidelity case over ten instances in terms of the SMSE and SMLL criteria.

It is observed that all the heterogeneous LMC models perform better than the simple \texttt{SOGP}, indicating successful knowledge transfer between the heterogeneous domains. Besides, the proposed \texttt{HSVLMC} model significantly outperforms the competitors in terms of both SMSE and SMLL. The superiority of \texttt{HSVLMC} in comparison to the competitors is mainly attributed to the flexible Bayesian calibration. 

\begin{figure*}[t!]
	\centering
	\includegraphics[width=.85\textwidth]{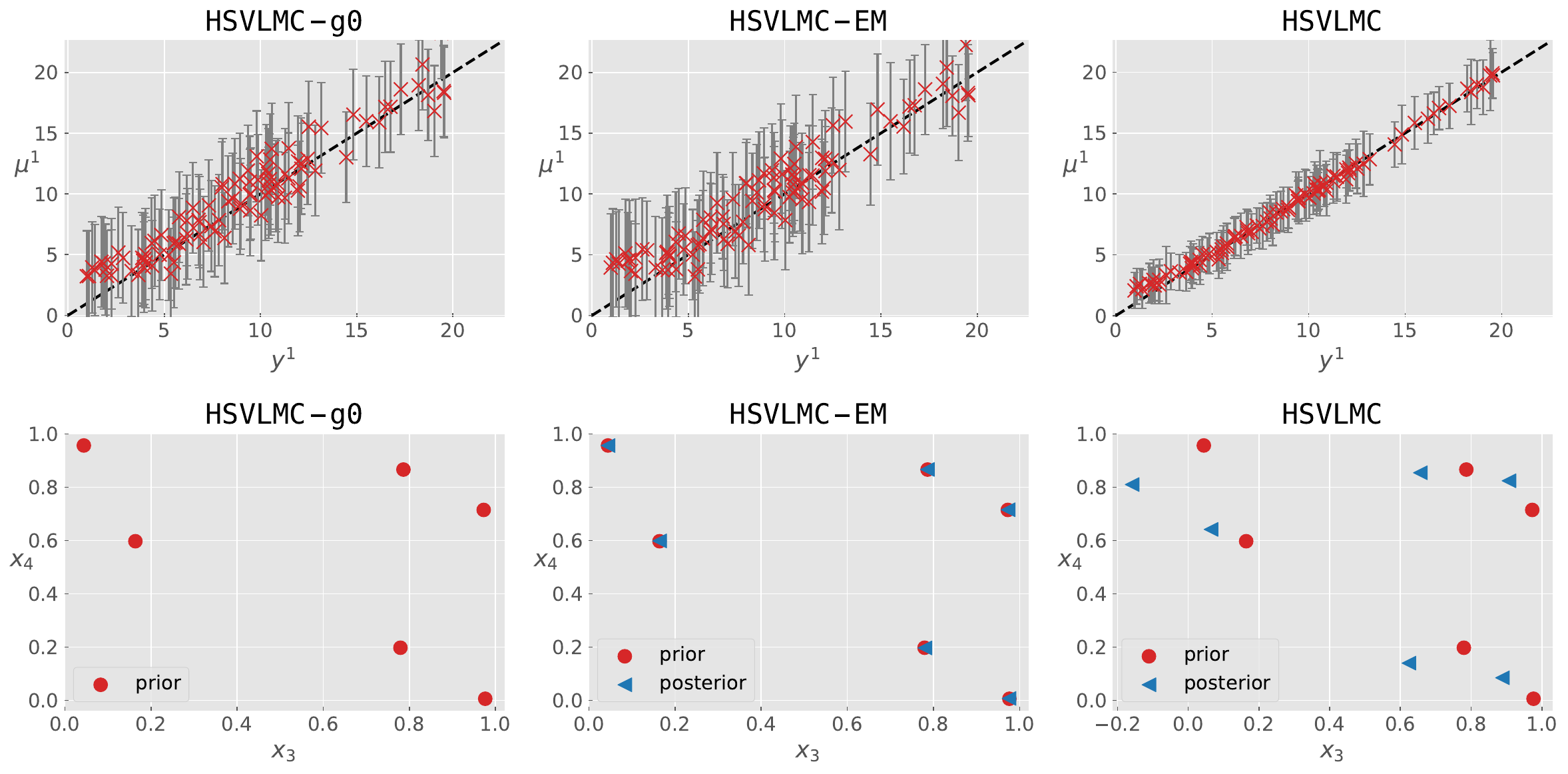}
	\caption{Predictions versus observations and the prior/posterior aligned inputs of different LMC models on the multi-fidelity case. Note that the crosses in the upper plots represent the prediction mean, while the error bars indicate 95\% confidence interval; the red circles in the bottom plots represent the means $\{\bm{\mu}_{g,i}^1\}$ of aligned inputs via the GP-inspired domain mapping $g^1(.)$, while the triangles represent the means $\{\bm{\mu}_{g_0,i}^1\}$ of aligned inputs via the prior domain mapping $g_0^1(.)$.}
	\label{fig_case2_corr_domain_mapping}
\end{figure*}

\begin{figure}[t!]
	\centering
	\includegraphics[width=.3\textwidth]{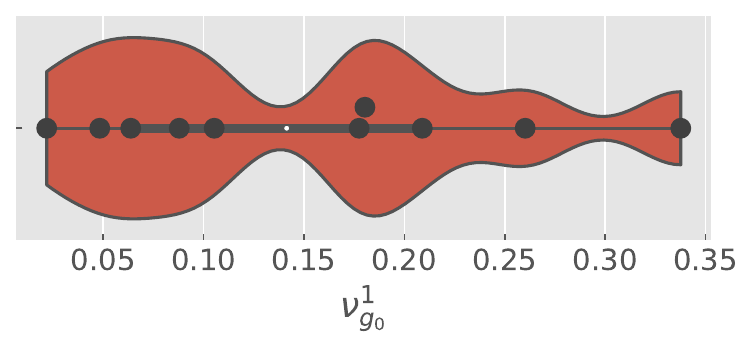}
	\caption{The violin plot of prior variance $\nu_{g_0}^1$ learned by the proposed \texttt{HSVLMC} on the multi-fidelity case over ten runs.}
	\label{fig_case2_var_g0}
\end{figure}

Fig.~\ref{fig_case2_corr_domain_mapping} further illustrates the diagonal plots of predictions versus observations and the comparison of the prior aligned inputs and the posterior aligned inputs. First, it is observed that the diagonal plots again verify the superiority of \texttt{HSVLMC} since its predictions agree well with the observations, and the estimated 95\% confidence interval covers the observations more compactly. Second, it is found that the \texttt{HSVLMC-g0} directly uses the prior aligned inputs; the aligned inputs learned by \texttt{HSVLMC-EM} almost covers the prior aligned inputs, since it attempts to fit the prior mapped inputs; differently, the aligned inputs learned by the proposed \texttt{HSVLMC} leave away from the prior inputs by taking into account the information loss brought by low-dimensional embedding, which in turn improves the quality of prediction. Finally, the superiority as well as the aligned inputs of \texttt{HSVLMC} is partially raised by the learned prior variance $\nu_{g_0}^1$. As illustrated in Fig.~\ref{fig_case2_var_g0}, it is observed that the prior variance is data dependent and thus varies over ten runs.

\subsection{Real-world datasets}
\subsubsection{Data description}
This section further verifies the performance of proposed \texttt{HSVLMC} model against competitors on five real-world heterogeneous multi-task cases, the summary of which is provided in Table~\ref{tab_real_dataset}.

\begin{table}
	\caption{Summary of five real-world datasets.} 
	\label{tab_real_dataset}
	\centering
		\begin{tabular}{lrrrr}
			\hline
			Datasets &$D^1$ &$D^2$ &$N^1$ &$N^2$ \\
			\hline
			$\mathtt{airfoil}$ &6 &3 &4 &1000 \\
			$\mathtt{disk}$ &6 &3 &8 &300 \\
			$\mathtt{supernova}$ &3 &2 &5 &1000 \\
			$\mathtt{sarcos^a}$ &21 &16 &20 &44484 \\
			$\mathtt{sarcos^b}$ &21 &16 &50 &50 \\
			\hline
		\end{tabular}
\end{table}

The first $\mathtt{airfoil}$ case is adapted from~\cite{zhu2011comparison} wherein the 52 input variables are employed to describe the airfoil geometry as well as the flight conditions including the speed and the angle of attack. This dataset has been further processed by using a dimensionality reduction algorithm like principle component analysis (PCA) to have six compressed inputs. Two different computational fluid dynamics (CFD) solvers with different fidelities are adopted to simulate the lift coefficient of airfoil. To construct the heterogeneous multi-task learning scenario, we perform Sobol sensitivity analysis~\cite{al2019meta} on the high-fidelity data to figure out the importance of inputs. Consequently, the low-fidelity task chooses the first, second and forth inputs as input parameters. That is, the prior domain mapping for this case is $g_0^1(x_1,x_2,x_3,x_4,x_5,x_6) = [x_1,x_2,x_4]^{\mathsf{T}}$. As described in Table~\ref{tab_real_dataset}, we randomly select $N^1=4$ six-dimensional high-fidelity points for task $y^1$ and $N^2=1000$ three-dimensional low-fidelity points for task $y^2$, and have 300 separate high-/low-fidelity points for testing.

The second $\mathtt{disk}$ case~\cite{zaytsev2016variable} uses six input variables to define the geometry of a rotating disk in an engine and outputs the maximum displacement. The case employs two finite element analysis (FEA) solvers with different fidelities to simulate the maximum displacement. Similarly, to construct the heterogeneous multi-task learning scenario, we perform Sobol sensitivity analysis on the high-fidelity data to select the first, fifth and sixth inputs for task $y^2$, thus resulting in the prior domain mapping for this case as $g_0^1(x_1,x_2,x_3,x_4,x_5,x_6) = [x_1,x_5,x_6]^{\mathsf{T}}$. As described in Table~\ref{tab_real_dataset}, we randomly select $N^1=8$ six-dimensional high-fidelity points for task $y^1$ and $N^2=300$ three-dimensional low-fidelity points for task $y^2$, and have 100 separate high-/low-fidelity points for testing.

The third $\mathtt{supernova}$ case comes from the Type Ia supernova red shift data on three cosmological physical constants including the Hubble constant, the dark matter and the dark energy fractions~\cite{davis2007scrutinizing}. The variable fidelity is performed by varying the grid size for a one-dimensional integration~\cite{zaytsev2016minimax}. Similarly, the Sobol sensitivity analysis helps select the first and second inputs for task $y^2$, thus resulting in the prior domain mapping for this case as $g_0^1(x_1,x_2,x_3) = [x_1,x_2]^{\mathsf{T}}$. As described in Table~\ref{tab_real_dataset}, we randomly select $N^1=5$ three-dimensional high-fidelity points for task $y^1$ and $N^2=1000$ two-dimensional low-fidelity points for task $y^2$, and have 499 separate high-/low-fidelity points for testing.

The final high-dimensional $\mathtt{sarcos}$ case comes from the inverse dynamic modeling of a seven-degree-of-freedom anthropomorphic robot arm. The twenty-one inputs of this case is composed of seven joint positions, seven joint velocities and seven joint accelerations, and the outputs are seven joint torques. To build the heterogeneous multi-task learning scenario, we use PCA to project the original twenty-one input variables into a sixteen-dimensional space. Besides, we build two cases ($\mathtt{sarcos^a}$ and $\mathtt{sarcos^b}$) from this dataset. The first \textit{asymmetric} $\mathtt{sarcos^a}$ case chooses the fourth torque with twenty-one inputs and the seventh torque with sixteen inputs, since the two torques in the original space has the highest spearman correlation of $r=0.96$. As described in Table~\ref{tab_real_dataset}, the $\mathtt{sarcos^a}$ case randomly select $N^1=20$ twenty-one-dimensional high-fidelity points for task $y^1$ and $N^2=44484$ sixteen-dimensional low-fidelity points for task $y^2$, and have 4449 separate high-/low-fidelity points for testing. Differently, the \textit{symmetric} $\mathtt{sarcos^b}$ case attempts to model the fourth torque with twenty-one inputs and the seventh torque with sixteen inputs jointly, with each torque having 50 training points and 4449 test points.

\subsubsection{Comparison results}
The numerical experiments on the five cases in Table~\ref{tab_real_dataset} are conducted over ten times, and the comparative results of different GP models in terms of both SMSE and SMLL are provided in Table~\ref{tab_real_dataset_comparison}. We have the following findings through the comparative results.

\begin{table*}
	\caption{Comparative results of different GP models on five heterogeneous multi-task learning cases in terms of both SMSE and SMLL. Note that the best results are marked in gray on each case.} 
	\label{tab_real_dataset_comparison}
	\centering
		\begin{tabular}{lrrrrr}
			\hline
			~ &Models &\texttt{SOGP} &\texttt{HSVLMC-g0} &\texttt{HSVLMC-EM} &\texttt{HSVLMC} \\
			\hline
			\multirow{5}*{SMSE} &$\mathtt{airfoil}$ &1.1329$_{\pm0.7121}$ &0.4301$_{\pm0.4112}$ &1.4758$_{\pm0.8637}$ &\cellcolor{mygray}0.1265$_{\pm0.2132}$ \\
			~ &$\mathtt{disk}$ &0.9068$_{\pm0.3325}$ &0.1702$_{\pm0.2485}$ &0.3066$_{\pm0.2691}$ &\cellcolor{mygray}0.0326$_{\pm0.0204}$ \\
			~ &$\mathtt{supernova}$ &0.7303$_{\pm0.3721}$ &0.5017$_{\pm0.4630}$ &0.6134$_{\pm0.5034}$ &\cellcolor{mygray}0.2147$_{\pm0.1339}$ \\
			~ &$\mathtt{sarcos^a}$ &0.3660$_{\pm0.1021}$ &0.1301$_{\pm0.0679}$ &0.3710$_{\pm0.1085}$ &\cellcolor{mygray}0.0580$_{\pm0.0134}$ \\
			~ &$\mathtt{sarcos^b}$ &0.1957$_{\pm0.1038}$ &0.1362$_{\pm0.0398}$ &0.1682$_{\pm0.0589}$ &\cellcolor{mygray}0.1217$_{\pm0.0257}$ \\
			\hline
			\hline
			\multirow{5}*{SMLL} &$\mathtt{airfoil}$ &295.3554$_{\pm758.1167}$ &0.7609$_{\pm3.1696}$ &0.7002$_{\pm1.0001}$ &\cellcolor{mygray}-0.7208$_{\pm1.2342}$ \\
			~ &$\mathtt{disk}$ &16.2522$_{\pm17.1235}$ &-0.9640$_{\pm0.8553}$ &-0.8760$_{\pm0.6581}$ &\cellcolor{mygray}-1.3207$_{\pm0.7411}$ \\
			~ &$\mathtt{supernova}$ &21.5982$_{\pm24.2586}$ &0.1099$_{\pm0.8263}$ &-0.0583$_{\pm0.6871}$ &\cellcolor{mygray}-0.4508$_{\pm0.9742}$ \\
			~ &$\mathtt{sarcos^a}$ &26.2674$_{\pm31.9417}$ &-1.0194$_{\pm0.2643}$ &-0.5825$_{\pm0.1908}$ &\cellcolor{mygray}-1.2654$_{\pm0.2140}$ \\
			~ &$\mathtt{sarcos^b}$ &2.8495$_{\pm1.6491}$ &-1.0171$_{\pm0.2531}$ &-0.8745$_{\pm0.3449}$ &\cellcolor{mygray}-1.1293$_{\pm0.1817}$ \\
			\hline
		\end{tabular}
\end{table*}

\textbf{The \texttt{SOGP} usually fails using a few training points.} For the asymmetric cases, since the target task $y^1$ only has a few number of training points, the \texttt{SOGP} cannot well predict at unseen points, indicated by the worst performance, especially the poor SMLL results, in Table~\ref{tab_real_dataset_comparison}. Contrarily, the heterogeneous LMC models leverage information from the related low-dimensional task, hence greatly improving the quality of prediction for the target task $y^1$ in comparison to the simple \texttt{SOGP}. Besides, it is observed that even for the symmetric $\mathtt{sarcos^b}$ case in Table~\ref{tab_real_dataset_comparison}, these heterogeneous LMC models outperform the \texttt{SOGP} by sharing knowledge across tasks.

\textbf{The \texttt{HSVLMC-g0} outperforms the \texttt{HSVLMC-EM} in general.} Compared to the direct usage of prior domain mapping in \texttt{HSVLMC-g0}, the \texttt{HSVLMC-EM} introduces an additional GP to fit it, which however may perform poorly when only a few number of training points are available. Hence, it is observed that the \texttt{HSVLMC-EM} even performs worse than the simple \texttt{SOGP} on some cases, for example, the $\mathtt{airfoil}$ case. But the Bayesian view of prior domain mapping may bring benefits for the \texttt{HSVLMC-EM} in some circumstances. For instance, it performs better than the \texttt{HSVLMC-g0} in terms of SMLL on two out of the five cases. Besides, the advantage of \texttt{HSVLMC-EM} in comparison to other LMC models is that it only requires knowing the aligned inputs $\bar{\mbf{X}}^1$ at training points, not the complete $g_0^1(.)$~\cite{hebbal2021multi}.

\textbf{The proposed \texttt{HSVLMC} significantly outperforms the others.} The superiority of the proposed \texttt{HSVLMC} over the others in terms of both SMSE and SMLL on all the cases is significant in Table~\ref{tab_real_dataset_comparison}. In comparison to the \texttt{HSVLMC-g0}, the Bayesian calibration in \texttt{HSVLMC} improves the model flexibility and capability by considering the effect of input alignment, which has been graphically illustrated in Fig.~\ref{fig_case2_corr_domain_mapping}. In comparison to the \texttt{HSVLMC-EM}, the \texttt{HSVLMC} do not directly model the prior domain mapping. Instead, it incorporates the KL divergence to penalize the discrepancy between the posterior mapping $q(\bar{\mbf{X}})$ and the prior mapping $p(\bar{\mbf{X}})$; meanwhile, it leaves flexibility for the adaption of aligned inputs, which contributes to the performance improvement.

\subsubsection{Discussions}
This section attempts to dive into the remarkable performance of the proposed \texttt{HSVLMC} model by assessing it in broader scenarios. That is, we seek to investigate the impact of training size $N^1$, dimensionality $D^2$, and task correlation on the model performance.

\begin{figure*}[t!]
	\centering
	\includegraphics[width=.65\textwidth]{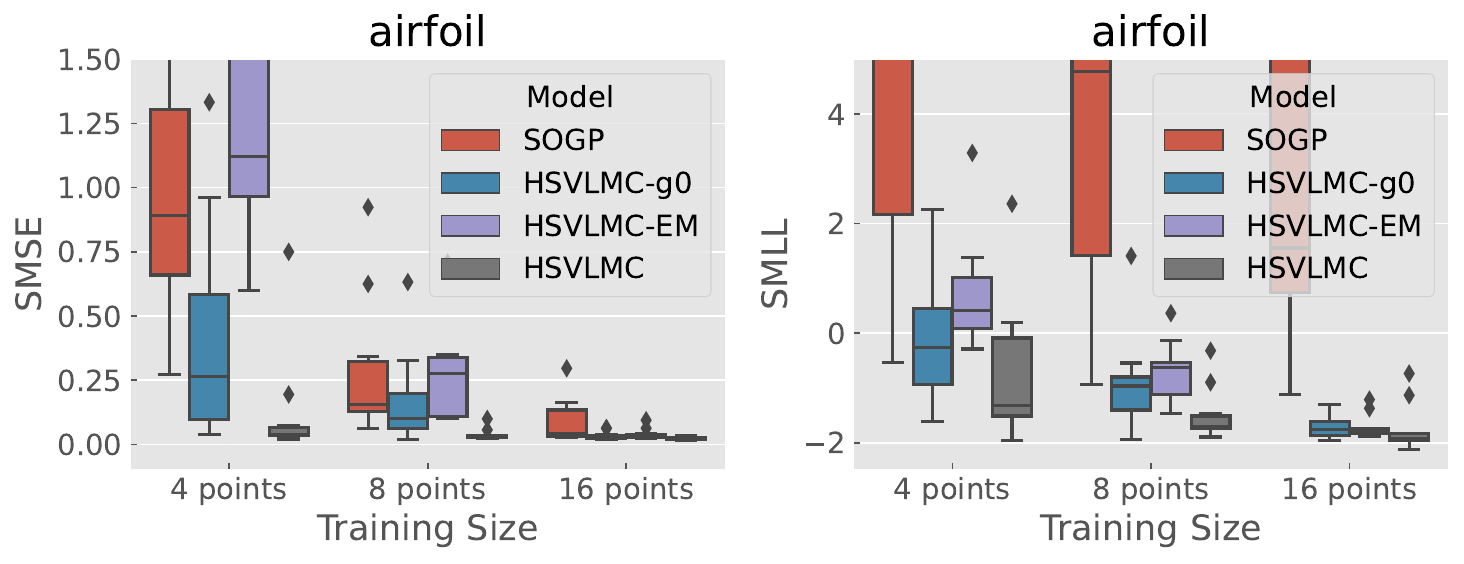}
	\caption{Impact of training size $N^1$ for the target task on the performance of heterogeneous LMC models on the $\mathtt{airfoil}$ case.}
	\label{fig_airfoil_N1}
\end{figure*}

Firstly, we investigate the impact of training size $N^1$ for the target task on the performance of heterogeneous LMC models in asymmetric scenario. To this end, Fig.~\ref{fig_airfoil_N1} depicts the comparative results on the $\mathtt{airfoil}$ case using $N^1=4$, $N^1=8$ and $N^1=16$, respectively. It is observed that with the increase of $N^1$, the performance of all the models has been improved, and the superiority of \texttt{HSVLMC-g0} and \texttt{HSVLMC} over the simple \texttt{SOGP} is always maintained. But this advantage gradually vanishes with further increase of $N^1$, e.g., the medium of predictions of \texttt{SOGP} is close to that of LMCs when $N^1=16$.

\begin{figure*}[t!]
	\centering
	\includegraphics[width=.65\textwidth]{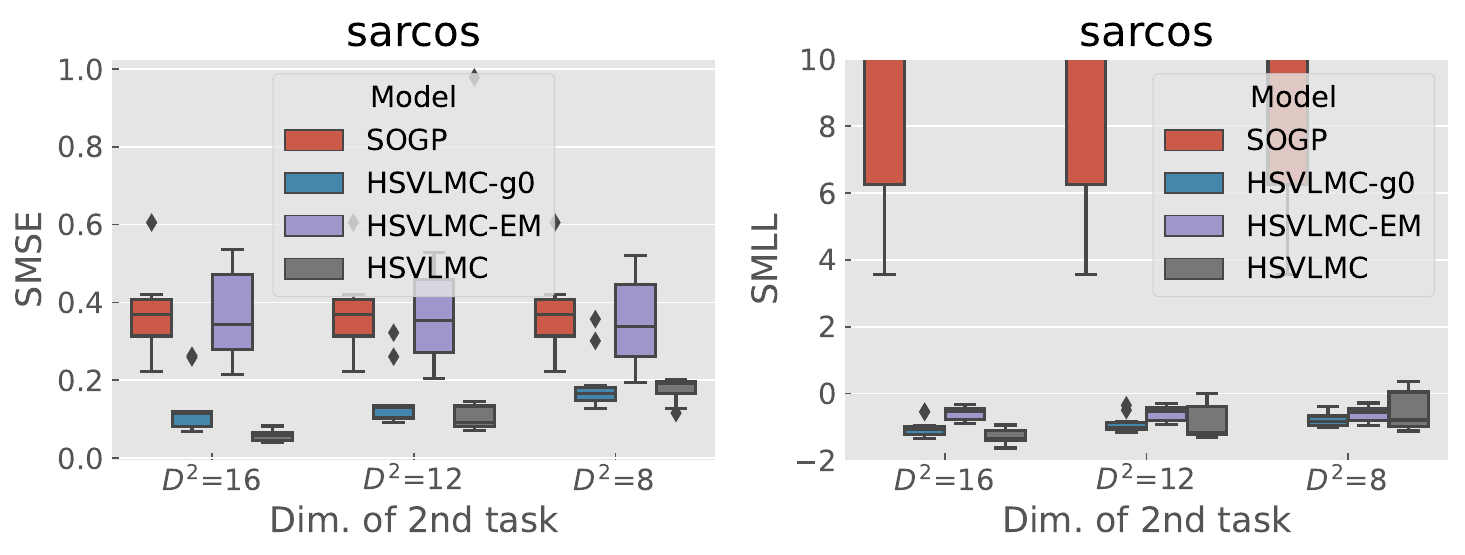}
	\caption{Impact of dimensionality $D^2$ of task $y^2$ on the performance of heterogeneous LMC models on the $\mathtt{sarcos^a}$ case.}
	\label{fig_sarcos_D2}
\end{figure*}

Secondly, we study the impact of dimensionality $D^2$ of task $y^2$ on the performance of heterogeneous LMC models. To this end, Fig.~\ref{fig_sarcos_D2} depicts the comparative results on the $\mathtt{sarcos^a}$ case using $D^2=16$, $D^2=12$ and $D^2=8$, respectively. With the decrease of dimensionality $D^2$, the information loss due to dimensionality reduction increases, thus increasing the difficulty of heterogeneous multi-task modeling. It is thus observed in Fig.~\ref{fig_sarcos_D2} that the performance of LMC models deteriorate with the decrease of $D^2$. Besides, the large dimensionality reduction from $D^1=21$ to $D^2=8$ makes the modeling more difficult such that the proposed \texttt{HSVLMC} performs slightly worse than the \texttt{HSVLMC-g0}, especially in terms of the SMLL results. But interestingly, the \texttt{HSVLMC-EM} and \texttt{HSVLMC} significantly outperform the \texttt{SOGP} even for the case with $D^2=8$.

\begin{figure*}[t!]
	\centering
	\includegraphics[width=.65\textwidth]{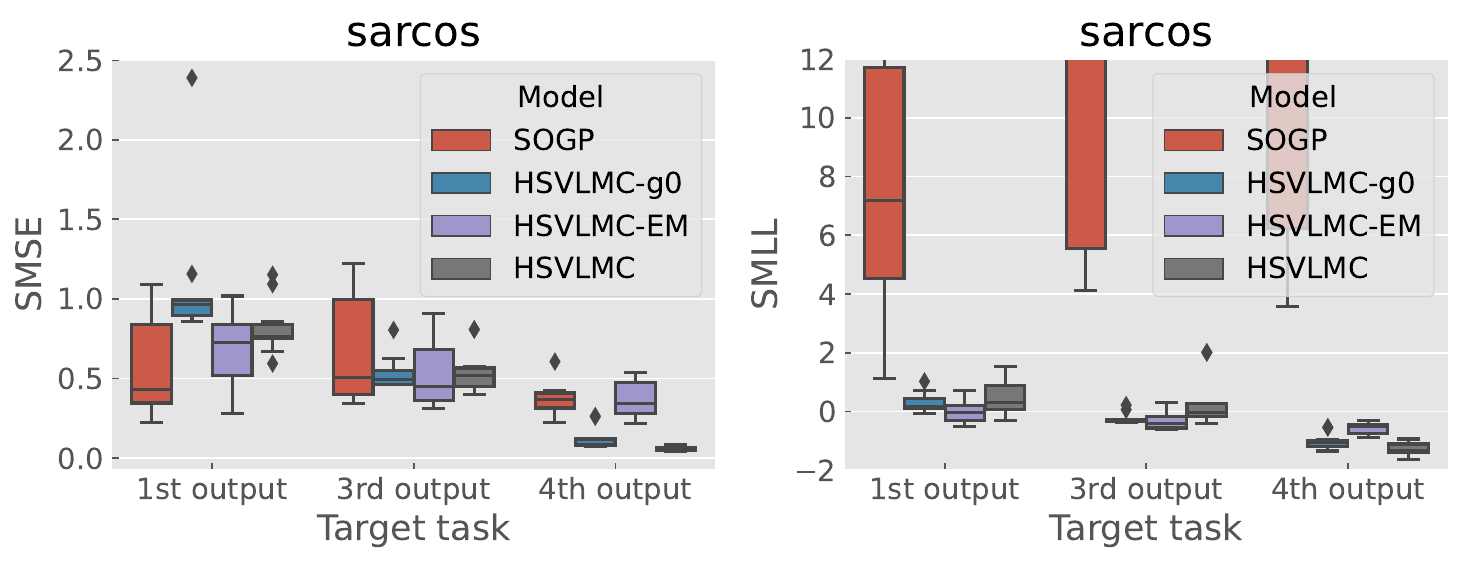}
	\caption{Impact of task correlation on the performance of heterogeneous LMC models on the $\mathtt{sarcos^a}$ case.}
	\label{fig_sarcos_corr}
\end{figure*}

Finally, the correlation of tasks may also affect the model performance. Therefore, Fig.~\ref{fig_sarcos_corr} investigates it by choosing different outputs as the first task on the $\mathtt{sarcos^a}$ case. It is found that the spearman correlation between the first and seventh outputs is only $r=0.41$; while this correlation between the third and seventh outputs is up to $r=0.74$; and as has been described before, the forth and seventh outputs have the highest correlation of $r=0.96$. Therefore, it is observed in Fig.~\ref{fig_sarcos_corr} that when the tasks are lowly correlated, the knowledge transferred from the low-dimensional task may deteriorate the prediction of LMC models in terms of SMSE. But with the increase of task correlation, the SMSE performance of LMC models gradually catches up with and even outperforms \texttt{SOGP}. Besides, the predictive distributions estimated by the LMCs better cover the data than the \texttt{SOGP} on this case, even when the tasks are lowly correlated. This is because without additional information, the \texttt{SOGP} tends to provide small uncertainty estimations given a few number of training points.

\section{Multi-fidelity modeling for steam turbine exhaust with heterogeneous inputs}
As an important component in the low-pressure steam turbines, the exhaust casing comprises the diffusing part to perform static pressure recovery, which contributes heavily to the overall effectiveness~\cite{yoon2011three}. The design and optimization of exhaust usually rely on expensive and time-consuming computational fluid dynamics (CFD) simulations. Therefore, data-driven surrogates or machine learning models have been employed to approximate and replace the expensive CFD simulator in order to speed up the downstream tasks~\cite{cremanns2018steam}. We here explore the application of the proposed \texttt{HSVLMC} to multi-fidelity modeling of the aerodynamic performance of the $\mathtt{exhaust}$ over heterogeneous inputs, the paradigm of which could further alleviate the computational budget as well as improving the flexibility. 

\begin{figure}[t!]
	\centering
	\includegraphics[width=.3\textwidth]{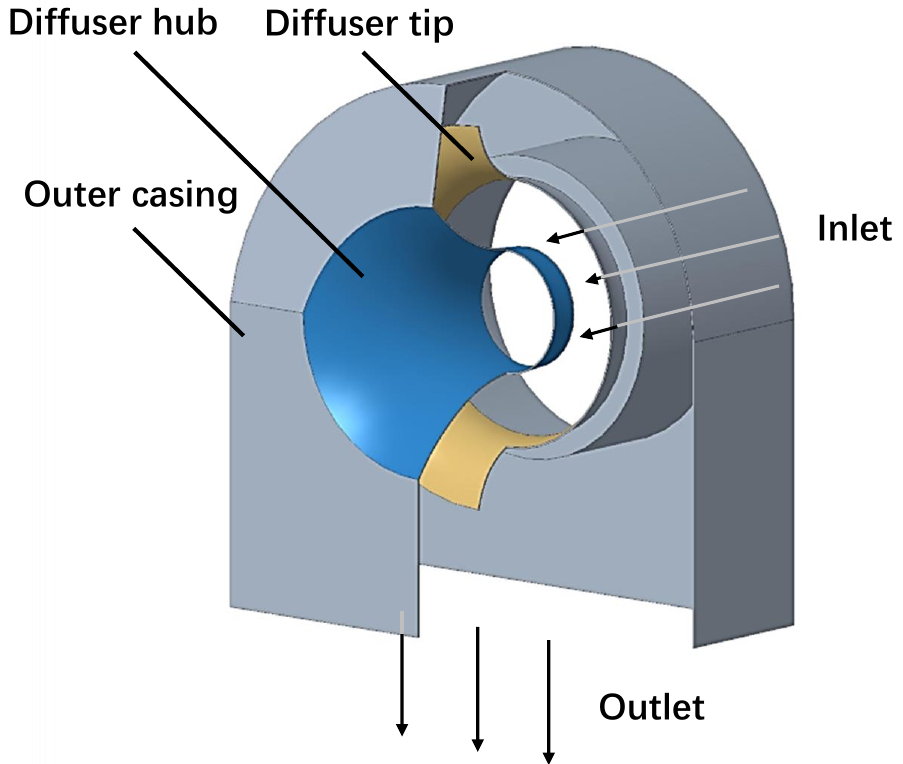}
	\caption{The geometry of studied low-pressure steam turbine exhaust.}
	\label{fig_exhaust}
\end{figure}

Fig~\ref{fig_exhaust} depicts the geometry of studied low-pressure steam turbine exhaust, and we particularly focus on the design of diffuser since it contributes mostly to the capability of static pressure recovery. To measure the aerodynamic performance of exhaust, we employ the total pressure loss coefficient
\begin{align}
	\zeta = \frac{P_{1t} - P_{2t}}{\frac{1}{2} \rho_1 \upsilon_1^2},
\end{align}
where $P_{1t}$ is the total pressure of inlet, $P_{2t}$ is the total pressure of outlet, $\rho_1$ is the density of inlet steam, and finally, $\upsilon_1$ is the velocity of inlet steam.

\begin{figure*}[t!]
	\centering
	\includegraphics[width=.7\textwidth]{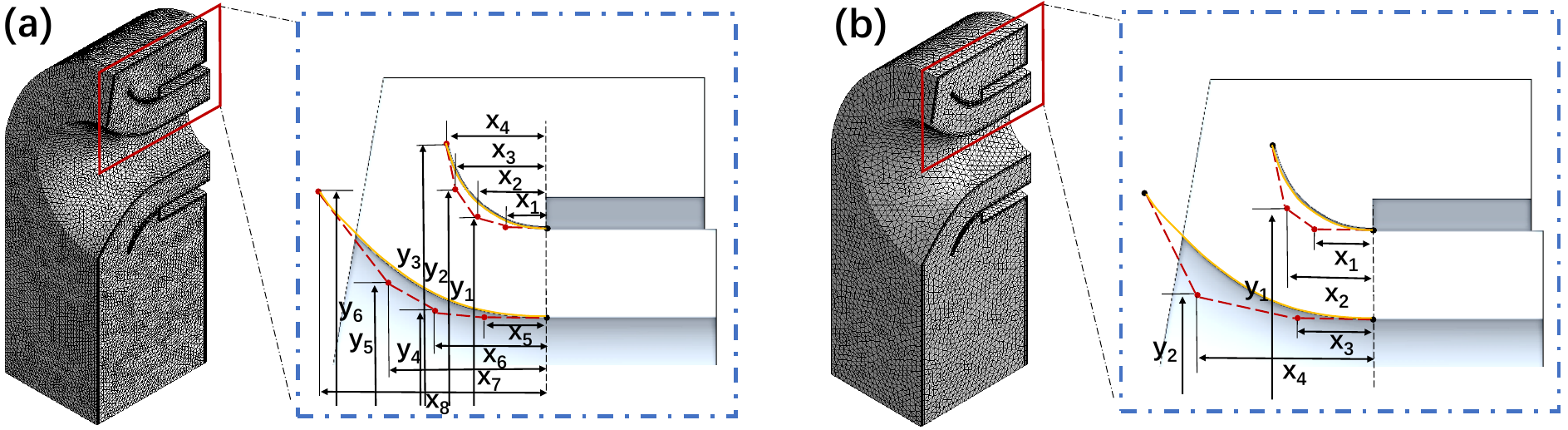}
	\caption{The (a) high-fidelity and high-dimensional $\mathtt{exhaust}$ model and the (b) low-fidelity and low-dimensional $\mathtt{exhaust}$ model. Note that because of the geometric symmetry, we take only half of the exhaust as computational domain.}
	\label{fig_exhaust_hf_lf}
\end{figure*}

We have two tasks for the $\mathtt{exhaust}$ problem. As shown in Fig.~{\ref{fig_exhaust_hf_lf}(a)}, the high-fidelity task parameterizes the diffuser tip and hub via B\'{e}zier curves with fourteen coordinate parameters, and performs a fine CFD simulation with 894756 unstructured meshes to obtain the coefficient $\zeta$. Contrarily, as shown in Fig.~{\ref{fig_exhaust_hf_lf}(b)}, the low-fidelity task parameterizes the diffuser tip and hub via B\'{e}zier curves with only six out of the fourteen coordinate parameters, and performs a coarse CFD simulation with 288439 unstructured meshes. The CFD simulation is conducted by the ANSYS CFX solver, and the configurations are listed in Table~\ref{tab_exaust_cfd}. It is found that a single high-fidelity simulation requires 0.5 hours on a personal computer with Intel i5 CPU and 8GB RAM, while a single low-fidelity CFD simulation only requires around 10 minutes. For this $\mathtt{exhaust}$ case, we generate only $N^1=10$ high-fidelity points and $N^2=200$ low-fidelity points for model training, and have separate 20 high-fidelity points for testing. 

\begin{table}
	\caption{The CFD configurations of steam turbine exhaust} 
	\label{tab_exaust_cfd}
	\centering
		\begin{tabular}{lr}
			\hline
			Parameters &Value \\
			\hline
			Inlet boundary &Mass flow $\dot{m}_1=74.736$kg/s \\
			& Total temperature $T_{1t}=336.15$K \\
			& Mass fraction of liquid phase is 0.1 \\
			Outlet boundary &Static pressure $P_2=6000$Pa \\
			Wall boundary &No slip \\
			Steam property & steam3vl \\
			Turbulent model &$k$-$\epsilon$ model \\
			Advection scheme &Upwind \\
			Turbulence numerics &First-order \\
			Algorithm &SIMPLE \\
			Convergence criteria &RMS less than $1e^{-5}$ \\
			\hline
		\end{tabular}
\end{table}

\begin{table}
	\caption{Comparative results of different models on the multi-fidelity $\mathtt{exhaust}$ problem with heterogeneous inputs in terms of both SMSE and SMLL. Note that the best results are remarked in gray.} 
	\label{tab_exaust}
	\centering
		\begin{tabular}{lrr}
			\hline
			Models &SMSE &SMLL \\
			\hline
			\texttt{SOGP} &1.2104$_{\pm0.2697}$ &21.8607$_{\pm49.4190}$ \\
			\texttt{HSVLMC-g0} &0.7090$_{\pm0.1777}$ &0.0298$_{\pm0.3725}$ \\
			\texttt{HSVLMC-EM} &0.7085$_{\pm0.1474}$ &-0.1286$_{\pm0.0984}$ \\
			\texttt{HSVLMC} &\cellcolor{mygray}0.5163$_{\pm0.2227}$ &\cellcolor{mygray}-0.2281$_{\pm0.3299}$ \\
			\hline
		\end{tabular}
\end{table}

Table~\ref{tab_exaust} reports the comparative results of different GP models on the multi-fidelity $\mathtt{exhaust}$ case with heterogeneous inputs. We observe that the proposed \texttt{HSVLMC} again outperforms the counterparts in terms of both SMSE and SMLL: it successfully transfers knowledge from the low-fidelity and low-dimensional data to help significantly improve the predictions with only 10 high-fidelity points in the high-dimensional space. As for the remaining two heterogeneous GPs, they also perform better than the simple \texttt{SOGP}, which fails (indicated by the poor SMSE and SMLL results) using only 10 high-fidelity and high-dimensional points. This practical engineering case highlights the benefits brought by our \texttt{HSVLMC} model: it bridges the gap among domains with heterogeneous inputs, and allows modeling and transferring knowledge from low-dimensional and relatively inexpensive domains.

\section{Conclusion}
\label{sec_conclusion}
To extend the LMC model, a well-known MTGP framework, for tackling heterogeneous multi-task cases, this paper presents the Bayesian calibration method to achieve input alignment. This method takes into account the effect of dimensionality reduction in order to learn good input alignment for the following multi-task modeling. Besides, it utilizes the residual mean in the posterior domain mappings to consider the inductive bias brought by the prior domain mappings. Consequently, the extensive numerical experiments in different heterogeneous multi-task scenarios have demonstrated the superiority of the proposed \texttt{HSVLMC} model. Further extensions would consider the challenging scenarios without prior domain mappings, and the application to downstream tasks, for example, multi-fidelity Bayesian optimization and evolutionary multi-tasking optimization over heterogeneous input domains.

\appendices
\section{Model configurations}
\label{app_confg}
This section provides the detailed model configurations for the comparative study in section~\ref{sec_exp}. First, for data preprocessing, we normalize the inputs and outputs along each dimension to normal distribution. We then randomly select the training and testing data according to the deployment in Table~\ref{tab_real_dataset}, and have ten instances to output convincing results.

Second, for the deployment of the proposed \texttt{HSVLMC} model, we use $Q=2$ latent GPs and employ the squared exponential kernel for each GP. For the kernel parameters, the output scale is initialized as 1.0; the lengthscales are initialized as 1.0 for the two $\mathtt{sarcos}$ cases and the modeling of $\mathtt{exhaust}$ problem, and are initialized as 0.1 for the remaining cases. Besides, we choose the same inducing size (i.e., $M_q=M$) for each of the $Q$ latent GPs, and take $M=30$ for the two toy cases, $M=200$ for the $\mathtt{exhaust}$ problem, and $M=100$ for the remaining cases. We initialize the positions of inducing points though the $k$-means clustering technique from the \texttt{scikit}-\texttt{learn} package~\cite{pedregosa2011scikit}. Particularly, for the MSGP in~\eqref{eq_msgp}, since the task $y^1$ usually has a few number of training points, we choose $M^1=N^1$ and initialize the inducing points as training points. Finally, for the prior variance $\nu_{g_0}^t$ in~\eqref{eq_px}, we initialize it as 0.1 for the $\mathtt{exhaust}$ problem, and take it as 1.0 for all the remaining cases.

Third, for the training of model, we employ the stochastic optimizer Adam ~\cite{kingma2015adam} and iteratively run it with a constant learning rate of $5\times10^{-3}$ over 5000 iterations. Particularly, for the large-scale $\mathtt{sarcos^a}$ and $\mathtt{sarcos^b}$ cases, we take the mini-batch size $|\mathcal{B}^2|=512$ for $y^2$ for efficient model training.

\bibliographystyle{IEEEtran}
\bibliography{IEEEabrv,HSVLMC}

\end{document}